\documentclass[10pt,twocolumn,letterpaper]{article}
\usepackage[pagenumbers]{cvpr} % arxiv-ready mode
\usepackage{graphicx}
\usepackage{amsmath}
\usepackage{amsfonts}
\usepackage{amssymb}
\usepackage{booktabs}
\usepackage{array}
\usepackage{algorithm}
\usepackage{algorithmic}
\usepackage{newfloat}
\usepackage{listings}
\DeclareCaptionStyle{ruled}{labelfont=normalfont,labelsep=colon,strut=off}
\floatstyle{ruled}
\newfloat{listing}{tb}{lst}{}
\floatname{listing}{Listing}
\definecolor{cvprblue}{rgb}{0.21,0.49,0.74}
\usepackage[pagebackref,breaklinks,colorlinks,allcolors=cvprblue]{hyperref}
\def\confName{CVPR}
\def\confYear{2027}
\AtBeginDocument{\bibliographystyle{ieeenat_fullname}}
\title{	
V-Zero: Gating On-Policy Distillation with Contrastive Visual Evidence for Fine-Grained Visual Reasoning}

\author{
Haoxiang Sun$^{1}$\thanks{Equal contribution.
\textsuperscript{\ensuremath{\dagger}}Corresponding author.},
Zhihang Yi$^{1*}$,
Langxuan Deng$^{1}$,
Yuhao Zhou$^{1}$,
Peiqi Jia$^{2}$,\\
Jian Zhao$^{3}$,
Li Yuan$^{4}$,
Jiancheng Lv$^{1}$,
Tao Wang$^{1,\dagger}$\\[4pt]
{\normalsize $^{1}$Sichuan University
\quad $^{2}$Xi'an Jiaotong University}
{\normalsize $^{3}$Northwest Polytechnical University
\quad $^{4}$Peking University}
}

\begin{document}

\maketitle

\begin{abstract}
Fine-grained visual reasoning requires multimodal large language models (MLLMs) to identify task-relevant visual evidence and ground their reasoning in local image regions. Existing methods typically rely on reinforcement learning with verifiable rewards or supervised fine-tuning on large-scale annotated reasoning traces, resulting in costly exploration, hand-designed verification rules, or heavy dependence on textual supervision. While On-Policy Distillation (OPD) offers a promising alternative via dense teacher supervision, its token-level objective inherently focuses on local token corrections, failing to assess global trajectory quality. In this work, we first reinterpret OPD as a stop-gradient alignment objective, revealing that its potential is limited by the absence of trajectory-level discrimination. Motivated by this observation, we propose V-Zero, which gates on-policy distillation with contrastive visual evidence. Specifically, V-Zero evaluates each trajectory under paired positive and negative visual views; the resulting likelihood contrast forms a trajectory-level gate that selectively amplifies updates from visually well-supported reasoning paths while suppressing ungrounded ones. Experiments on multiple visual reasoning and perception benchmarks show that V-Zero improves the Qwen3.5-4B backbone by 3.9 points on average while preserving strong generalization. Notably, V-Zero  achieves more than 3$\times$ and 6$\times$ training speedups over representative supervised fine-tuning and reinforcement learning baselines, respectively. It requires neither annotated textual reasoning traces nor external reward signals. Code and dataset are released at https://github.com/eVI-group-SCU/V-Zero.
% will be released upon acceptance.
\end{abstract}

\section{Introduction}

\begin{figure}[t]
    \centering
    \includegraphics[width=\linewidth]{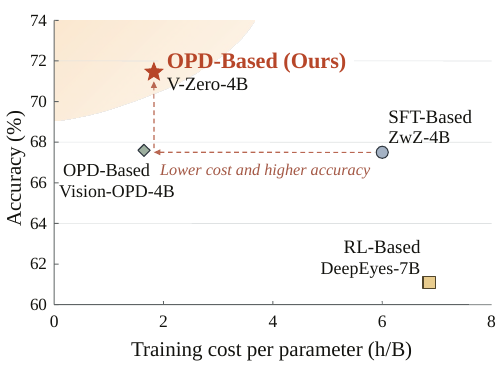}
    \caption{Accuracy versus training cost per parameter, measured in hours per billion parameters (h/B). V-Zero achieves higher accuracy at lower cost than ZwZ and DeepEyes. V-Zero also outperforms Vision-OPD by 3.9 percentage points in accuracy  with only one additional teacher forward pass during training.}
    \label{fig:paradigms}
\end{figure}

As Multimodal Large Language Models (MLLMs) rapidly develop~\cite{bai2025qwen3,comanici2025gemini}, fine-grained visual reasoning~\cite{wu2024v,hrbench} has become a critical capability for evaluating them. Unlike general visual understanding~\cite{yu2023mm,yue2024mmmu,liu2024mmbench}, fine-grained visual reasoning requires models to inspect local details, identify task-relevant visual evidence, and reason over specific image regions.

Recent studies have explored the integration of agentic visual search and reasoning~\cite{zheng2026deepeyesincentivizingthinkingimages,zhang2025thymethinkimages}, often referred to as \textit{thinking with images}~\cite{su2025thinkingimagesmultimodalreasoning}. By interleaving reasoning with visual search, this paradigm enables models to decide where to look, gather task-relevant visual evidence, and refine their answers in a grounded manner. Despite their promise, these methods~\cite{zheng2026deepeyesincentivizingthinkingimages,zhang2025thymethinkimages} often rely on Reinforcement Learning (RL), which incurs costly exploration and requires predefined verifiable rules for training signals. Another line of work~\cite{wei2026zoomingzoomingregiontoimagedistillation} adopts Supervised Fine-tuning (SFT) on large-scale annotated image-text data, achieving promising results but requiring massive textual supervision and risking catastrophic forgetting~\cite{chu2025sft}.

On-Policy Distillation (OPD)
provides a promising alternative to SFT and RL by supplying dense supervision
on trajectories sampled from the student itself. However, standard OPD
provides only local supervision at each student-induced prefix. Once the
student drifts onto an erroneous reasoning path, the teacher continues to
provide next-token targets conditioned on that prefix, without a
trajectory-level signal to distinguish or suppress the drifting
path~\cite{fu2026revisiting}. These observations motivate the central question
of this work:

\textbf{\textit{
How can efficient post-training move beyond local token-
level correction to selectively reinforce reliable, visually
grounded reasoning trajectories, without relying on costly
RL exploration or large-scale text supervision?}}

In this paper, we first develop a complementary view of OPD by reinterpreting it as a stop-gradient alignment objective: at each student-generated prefix, the student matches a fixed teacher target for the next token. This perspective explains why OPD provides effective dense on-policy supervision while revealing its lack of explicit trajectory-level discrimination for erroneously drifting trajectories. Building on this perspective, we propose V-Zero, which introduces a trajectory-level evidence gate derived from contrastive visual evidence. For each student trajectory, the teacher evaluates the same token sequence under a positive view cropped from the task-relevant region and a negative view cropped from an irrelevant random region. The resulting trajectory-level likelihood contrasts are compared and normalized within each rollout group to construct the gate, which upweights distillation for reasoning paths with stronger relative visual support while suppressing weakly supported ones.

Notably, V-Zero requires neither annotated textual reasoning traces nor external reward signals. Experiments across multiple visual reasoning and perception benchmarks show that it improves the Qwen3.5-4B by 3.9 points on average while preserving strong generalization. Moreover, V-Zero achieves more than 3$\times$ and 6$\times$ training speedups over representative supervised fine-tuning and reinforcement learning baselines, respectively. These gains are obtained using contrastive visual evidence only during training, without introducing additional tool calls or computational overhead at inference time.

In summary, our contributions are as follows:
\begin{itemize}
    \item \textbf{A theoretical view of OPD.}
    Through the lens of contrastive learning, we reinterpret OPD as a stop-gradient alignment  between student and teacher views, revealing its lack of explicit trajectory-level discrimination.

    \item \textbf{Contrastive evidence gating mechanism.}
    We propose V-Zero, which derives a trajectory-level gate from paired positive and negative visual evidence views to guide trajectory-level distillation during OPD training.

    \item \textbf{Efficient and generalizable visual reasoning.}
    Extensive experiments show that the proposed gate improves fine-grained visual reasoning while preserving general multimodal capabilities and standard inference-time behavior.
\end{itemize}

\section{Related Work}

\noindent\textbf{Agentic Visual Reasoning.}
Fine-grained multimodal reasoning requires models to identify and use small
but critical visual evidence. Standard MLLMs struggle when answers depend on
localized visual search rather than global scene understanding~\cite{wu2024v,hrbench}.
Recent works address this limitation by training MLLMs to interleave reasoning
with visual operations, allowing models to gather new visual observations during
inference~\cite{zheng2026deepeyesincentivizingthinkingimages,wang2025pixelreasonerincentivizingpixelspace,fan2025gritteachingmllmsthink,zhang2025thymethinkimages}.
However, these methods typically require costly RL exploration, predefined
verifiable rewards, and additional inference-time operations. ZwZ~\cite{wei2026zoomingzoomingregiontoimagedistillation}
shows that comparable performance can be achieved without RL by scaling
supervised fine-tuning, but it requires large-scale annotated image-text data
and increases the risk of catastrophic forgetting in MLLMs.

\noindent\textbf{On-Policy Distillation.}
OPD trains on trajectories sampled from the student itself and uses a teacher to
provide dense supervision on student-induced states~\cite{agarwal2024onpolicydistillationlanguagemodels,lu2025onpolicydistillation}.
Recent studies show that OPD can serve as an efficient post-training recipe,
mitigating catastrophic forgetting while converging quickly~\cite{li2026rethinkingonpolicydistillationlarge,shenfeld2026selfdistillationenablescontinuallearning}.
Other works extend OPD to self-distillation settings, where teacher and student
are constructed from the same model under different conditions~\cite{zhao2026selfdistilledreasoneronpolicyselfdistillation,yang2026selfdistilledrlvr},
or combine it with reinforcement learning to provide dense learning signals
while preserving reward-based optimization for task correctness~\cite{hubotter2026reinforcement}.
In multimodal settings, Video-OPD~\cite{li2026videoopdefficientposttrainingmultimodal}
extends OPD to temporal video grounding and shows that teacher-provided
token-level supervision on on-policy trajectories can outperform GRPO with
faster convergence. Concurrent with our work, Vision-OPD~\cite{yuan2026vision},
performs regional-to-global self-distillation using a crop-conditioned teacher
and a full-image-conditioned student instantiated from the same MLLM.
Although its formulation is closely related to ours, Vision-OPD primarily
studies how OPD can be adapted to fine-grained vision tasks. In contrast, our
focus is on introducing a
trajectory-level evidence gate into OPD, using paired positive and negative
visual views to distinguish and selectively weight student-generated
trajectories.

\section{Revisiting OPD as Stop-Gradient Alignment without Negative Pairs}
\begin{figure*}[t]
    \centering
    \includegraphics[width=1.0\linewidth]{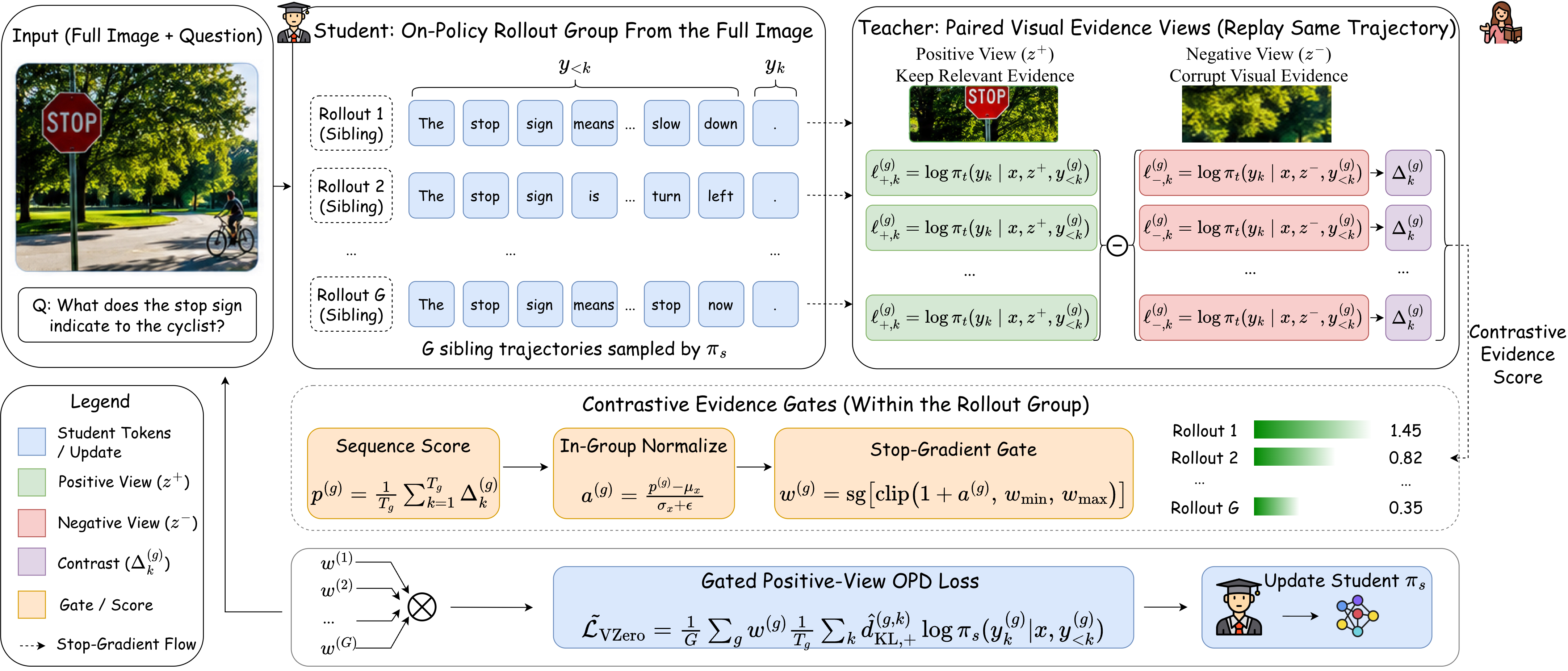}
    \caption{Overview of V-Zero. Sibling rollouts are parallel reasoning
    trajectories sampled for the same prompt. The student samples them from the
    full image, while a teacher-side evidence comparison module replays them
    under paired positive and negative visual evidence views to produce
    trajectory-level contrastive evidence gates. }
    \label{fig:method-overview}
\end{figure*}

Before presenting V-Zero, we revisit OPD as an alignment objective on student-induced states. OPD efficiently provides dense token-level correction by matching student predictions to teacher targets on sampled prefixes, but it lacks trajectory-level discriminative supervision.

\subsection{On-Policy Distillation with Teacher-Side Views}

OPD trains a student policy $\pi_s$ on states generated by the student itself. Let $\mathcal{D}=\{x_i\}_{i=1}^{N}$ be a
set of prompts. For each prompt $x$, the student samples a group of $G$
on-policy trajectories $\mathcal{Y}(x)=\{y^{(g)}\}_{g=1}^{G}$, with the standard
single-rollout case recovered when $G=1$. Each trajectory
$y^{(g)}=(y^{(g)}_1,\ldots,y^{(g)}_{T_g})$ is generated autoregressively as
\begin{equation}
    y^{(g)}_k \sim \pi_s(\cdot \mid x,y^{(g)}_{<k}),
    \quad g=1,\ldots,G,\quad k=1,\ldots,T_g.
\end{equation}
We denote the resulting group rollout distribution by $\pi_s^G(\cdot\mid x)$.
The sampled trajectories are treated as stop-gradient training data. The teacher is then queried on the same student-induced prefixes, and
the student is optimized to match the teacher on the states it actually visits.
Throughout the paper, RKL denotes the reverse Kullback--Leibler divergence
(reverse KL). We first define its local full-vocabulary form at token position
$k$ of rollout $g$:
\begin{equation}
    D_{\mathrm{KL}}^{(g,k)}
    =
    \sum_{v\in\mathcal{V}}
    \pi_s(v\mid x,y^{(g)}_{<k})
    \log
    \frac{
        \pi_s(v\mid x,y^{(g)}_{<k})
    }{
        \pi_t(v\mid x,y^{(g)}_{<k})
    } .
\end{equation}
The group-level OPD objective is then
\begin{equation}
    \mathcal{L}_{\mathrm{OPD}}^{\mathrm{RKL}}(\pi_s)
    =
    \mathbb{E}_{\substack{x\sim\mathcal{D},\ \mathcal{Y}(x)\sim\pi_s^G(\cdot\mid x)}}
    \left[
        \mathcal{L}_{\mathrm{OPD}}^{\mathrm{RKL}}(x,\mathcal{Y}(x))
    \right].
\end{equation}
\begin{equation}
    \mathcal{L}_{\mathrm{OPD}}^{\mathrm{RKL}}(x,\mathcal{Y}(x))
    =
    \frac{1}{G}\sum_{g=1}^{G}
    \frac{1}{T_g}
    \sum_{k=1}^{T_g}
    D_{\mathrm{KL}}^{(g,k)}.
\end{equation}
In practice, sampled-token OPD~\cite{lu2025onpolicydistillation,fu2026revisiting,li2026rethinkingonpolicydistillationlarge} is used to form a sampled
log-ratio score for this local reverse-KL objective. Here
$\mathrm{sg}[\cdot]$ denotes the stop-gradient operator:
\begin{align}
    \widehat{d}_{\mathrm{KL}}^{(g,k)}
    &=
    \mathrm{sg}\!\left[
    \log
    \frac{
        \pi_s(y^{(g)}_k\mid x,y^{(g)}_{<k})
    }{
        \pi_t(y^{(g)}_k\mid x,y^{(g)}_{<k})
    }
    \right].
\end{align}
\begin{align}
    y^{(g)}_k
    &\sim \pi_s(\cdot\mid x,y^{(g)}_{<k}).
\end{align}
To optimize this reverse-KL minimization objective with student-sampled tokens,
we use a stop-gradient sampled surrogate:
\begin{equation}
    \widetilde{\ell}_{\mathrm{OPD}}^{(g,k)}
    =
    \widehat{d}_{\mathrm{KL}}^{(g,k)}
    \log \pi_s(y^{(g)}_k\mid x,y^{(g)}_{<k}).
\end{equation}
This formulation naturally extends to training with privileged information. The
student still samples trajectories from the original prompt $x$, while the teacher
may condition on additional information $z$ that is unavailable to the student,
such as a localized crop or a reference solution~\cite{zhao2026selfdistilledreasoneronpolicyselfdistillation}. The teacher target is
then evaluated as
\begin{equation}
    \pi_t(\cdot \mid x,z,y^{(g)}_{<k}),
\end{equation}
and the OPD objective is obtained by replacing
$\pi_t(\cdot\mid x,y^{(g)}_{<k})$ with
$\pi_t(\cdot\mid x,z,y^{(g)}_{<k})$.

\begin{figure*}[t]
    \centering
    \includegraphics[width=1.0\textwidth]{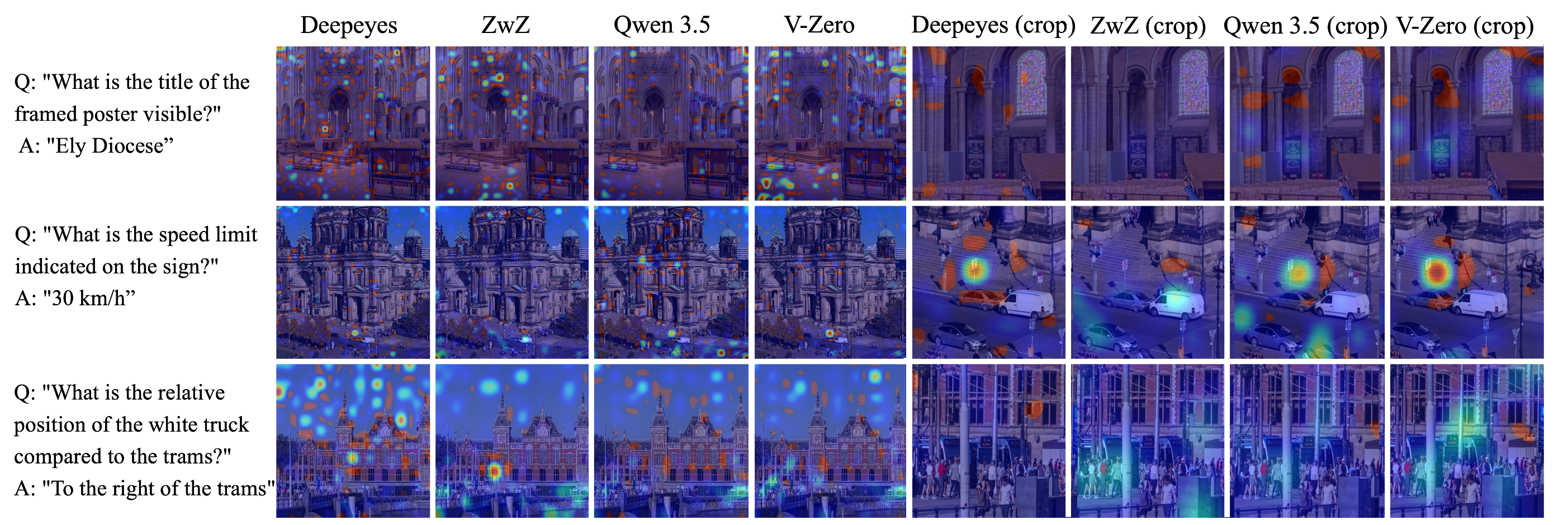}
    \caption{Attention visualization on representative fine-grained reasoning
    samples. In the first row, the question focuses on the title of the framed
    poster in the lower-right image region; V-Zero and the Qwen3.5-4B baseline are
    the only methods that cover the correct visual area, with V-Zero producing
    stronger activation. In the second row, the answer depends on the speed
    limit sign near the bottom of the image, where V-Zero shows the strongest
    focus. In the third row, the question requires the spatial relation between
    the white truck and the trams, and V-Zero is the only method that clearly
    highlights both visual targets.}
    \label{fig:placeholder-visualization}
\end{figure*}

\subsection{An Asymmetric Alignment View of OPD}

The privileged-information formulation reveals an asymmetric alignment structure underlying OPD.
For each student-induced state $(x,y^{(g)}_{<k})$, the student branch defines a
base view $v_s^{(g,k)}=(x,y^{(g)}_{<k})$, while the teacher branch defines a
target view $v_t^{(g,k)}$. In standard OPD the two views share the same context;
with teacher-side information, the teacher view is augmented to
$v_t^{(g,k)}=(x,z,y^{(g)}_{<k})$. These two views induce predictive
distributions over the same next-token decision:
\begin{equation}
    q_s^{(g,k)}
    =
    \pi_s(\cdot \mid v_s^{(g,k)}),
    \qquad
    q_t^{(g,k)}
    =
    \mathrm{sg}\!\left[
        \pi_t(\cdot \mid v_t^{(g,k)})
    \right].
\end{equation}
Here $\pi_s(\cdot \mid v_s^{(g,k)})$ abbreviates
$\pi_s(\cdot \mid x,y^{(g)}_{<k})$, and
$\pi_t(\cdot \mid v_t^{(g,k)})$ abbreviates either
$\pi_t(\cdot \mid x,y^{(g)}_{<k})$ in standard OPD or
$\pi_t(\cdot \mid x,z,y^{(g)}_{<k})$ when teacher-side information is used.
The stop-gradient operator makes the alignment asymmetric: the student remains
the online branch to be optimized, while the teacher provides a fixed target.

Thus, OPD can be viewed as a negative-free\footnote{We use negative-free in the sense of contrastive learning.}  stop-gradient alignment objective over student-teacher views:
\begin{equation}
    \ell_{\mathrm{align}}^{(g,k)}
    =
    d\!\left(q_s^{(g,k)}, q_t^{(g,k)}\right),
\end{equation}
where $d(\cdot,\cdot)$ can be instantiated by the sampled-token reverse-KL
score. The corresponding stop-gradient sampled score is
\begin{align}
    \widehat{d}_{\mathrm{KL,align}}^{(g,k)}
    &=
    \mathrm{sg}\!\left[
    \log q_s^{(g,k)}(y^{(g)}_k)
    -
    \log q_t^{(g,k)}(y^{(g)}_k)
    \right].
\end{align}
\begin{align}
    y^{(g)}_k
    &\sim q_s^{(g,k)}.
\end{align}
The corresponding surrogate loss is
\begin{equation}
    \widetilde{\ell}_{\mathrm{align}}^{(g,k)}
    =
    \widehat{d}_{\mathrm{KL,align}}^{(g,k)}
    \log q_s^{(g,k)}(y^{(g)}_k).
\end{equation}

This view also exposes a key limitation of standard OPD. Although OPD provides dense token-level alignment, it does not explicitly score the correctness of the full trajectory. Once the student enters an erroneous reasoning path, the teacher can only provide local next-token targets conditioned on that prefix, without assessing whether the trajectory as a whole is approaching the correct answer. As a result, standard OPD may optimize locally plausible continuations while lacking trajectory-level discriminative supervision. V-Zero addresses this limitation by estimating rollout reliability through paired positive and negative teacher-side visual evidence views and using trajectory-level contrastive evidence gates to modulate
dense token-level distillation.

\section{Method}
As illustrated in Figure~\ref{fig:method-overview}, V-Zero improves fine-grained visual reasoning by adding a contrastive evidence gating mechanism to on-policy distillation. The student samples on-policy trajectories from the full image, while the teacher replays the same trajectories with additional paired positive and negative visual evidence views. 

\noindent\textbf{Notation.}
For prompt $x_i$, $y_i^{(g)}$ is the $g$-th sibling rollout,
$y_{i,<k}^{(g)}$ is its prefix before token $k$, and $T_i^{(g)}$ is its length.
The positive and negative teacher-side evidence views are denoted by $z_i^+$
and $z_i^-$, respectively. We use $i$, $g$, and $k$ for prompt, rollout, and
token indices; when discussing a single prompt, the prompt index $i$ is
suppressed for readability.
\begin{algorithm}[t]
\caption{V-Zero Training}
\label{alg:vzero}
\textbf{Input}: dataset $\mathcal{D}$, student $\pi_s$, teacher $\pi_t$, group size $G$\\
\textbf{Hyperparameters}: $w_{\min}, w_{\max}$
\begin{algorithmic}[1]
\FOR{each training step}
    \STATE $\mathcal{B} \leftarrow$ sample minibatch from $\mathcal{D}$
    \FOR{each prompt $x_i \in \mathcal{B}$}
        \STATE $\{y_i^{(g)}\}_{g=1}^{G} \leftarrow$ sample $G$ rollouts from $\pi_s(\cdot \mid x_i)$
        \STATE $z_i^+ \leftarrow$ positive visual evidence view
        \STATE $z_i^- \leftarrow$ negative visual evidence view
        \FOR{$g=1,\ldots,G$}
            \STATE compute $\ell_{s,i,k}^{(g)}$ with $(x_i,y_{i,<k}^{(g)})$
            \STATE compute $\ell_{+,i,k}^{(g)}$ with $(x_i,z_i^+,y_{i,<k}^{(g)})$
            \STATE compute $\ell_{-,i,k}^{(g)}$ with $(x_i,z_i^-,y_{i,<k}^{(g)})$
            \STATE $\Delta_{i,k}^{(g)} \leftarrow \ell_{+,i,k}^{(g)}-\ell_{-,i,k}^{(g)}$
            \STATE $p_i^{(g)} \leftarrow \frac{1}{T_i^{(g)}}\sum_{k=1}^{T_i^{(g)}}\Delta_{i,k}^{(g)}$
        \ENDFOR
        \STATE $(\mu_i,\sigma_i) \leftarrow \operatorname{MeanStd}_{g=1}^{G}(p_i^{(g)})$
        \FOR{$g=1,\ldots,G$}
            \STATE $a_i^{(g)} \leftarrow \frac{p_i^{(g)}-\mu_i}{\sigma_i+\epsilon}$
            \STATE $w_i^{(g)} \leftarrow \mathrm{sg}\!\left[\mathrm{clip}(1+a_i^{(g)}, w_{\min}, w_{\max})\right]$
            \STATE $\widehat{d}_{\mathrm{KL},+,i,k}^{(g)} \leftarrow \mathrm{sg}\!\left[\ell_{s,i,k}^{(g)}-\ell_{+,i,k}^{(g)}\right]$ for all valid $k$
        \ENDFOR
    \ENDFOR
    \STATE compute each rollout surrogate $\widetilde{\ell}_i^{(g)}$ as in Eq.~\eqref{surrogate}
    \STATE $\widetilde{\mathcal{L}} \leftarrow
    \frac{1}{|\mathcal{B}|G}\sum_{i,g}\widetilde{\ell}_i^{(g)}$
    \STATE update $\pi_s$ using $\nabla \widetilde{\mathcal{L}}$
\ENDFOR
\end{algorithmic}
\end{algorithm}

\subsection{Student Rollouts and Teacher Evidence Views}

Given a prompt $x$ with the original full image, the student samples a group of
$G$ trajectories:
\begin{equation}
    \mathcal{Y}(x)=\{y^{(g)}\}_{g=1}^{G},
    \qquad
    y^{(g)} \sim \pi_s(\cdot \mid x).
\end{equation}
These trajectories are sibling rollouts from the same prompt and policy. For each sampled trajectory, the teacher replays the same token sequence with the original full image plus an additional pair of visual evidence views. The positive view $z^{+}$ is a target-region crop that preserves task-relevant visual evidence. For $z^{-}$, we downsample the original image by a factor of two and then randomly sample a crop with the same pixel dimensions as $z^+$, constrained to lie outside the target region. This teacher-side evidence comparison estimates how strongly each rollout depends on the relevant visual evidence. The teacher then computes sampled-token log-probabilities under the two additional views:
\begin{align}
    \ell_{+,k}^{(g)}
    &=
    \log \pi_t(y_k^{(g)} \mid x,z^{+},y_{<k}^{(g)}),
    \\
    \ell_{-,k}^{(g)}
    &=
    \log \pi_t(y_k^{(g)} \mid x,z^{-},y_{<k}^{(g)}).
\end{align}

The two teacher replays use identical response tokens, so the comparison holds the student's reasoning path fixed while changing the additional visual evidence. This makes the score sensitive to how the teacher's support for that particular path changes with the supplied region. The original image remains available as context in both replays, allowing the crop to emphasize local details without removing the surrounding scene.

\subsection{Contrastive Evidence Gating}

Given the positive and negative teacher evaluations above, V-Zero turns visual
dependence into a contrastive signal. Intuitively, tokens that genuinely rely on
task-relevant visual evidence should receive stronger teacher support from the
target-region crop than from the downsampled irrelevant region. For each
student-sampled token, we first compute the teacher-side visual evidence gap:
\begin{equation}
    \Delta_k^{(g)}
    =
    \ell_{+,k}^{(g)}-\ell_{-,k}^{(g)} .
\end{equation}
A larger $\Delta_k^{(g)}$ indicates that the token is more strongly supported
when the teacher has access to the relevant visual evidence. We then aggregate
these token-level gaps into a trajectory-level evidence score:
\begin{equation}
    p^{(g)}
    =
    \frac{1}{T_g}
    \sum_{k=1}^{T_g}
    \Delta_k^{(g)} .
\end{equation}
Since raw evidence scores can vary across prompts, answer lengths, and visual
contexts, V-Zero normalizes the sibling score vector
$\mathbf{p}(x)=(p^{(1)},\ldots,p^{(G)})$ within each prompt:
\begin{equation}
    (\mu_x,\sigma_x)
    =
    \operatorname{MeanStd}(\mathbf{p}(x)),
    \qquad
    a^{(g)}
    =
    \frac{p^{(g)}-\mu_x}{\sigma_x+\epsilon}.
\end{equation}
The normalized quantity $a^{(g)}$ is a trajectory-level evidence advantage: it
measures whether the current rollout is better visually grounded than its
siblings under the same prompt. V-Zero converts this advantage into a
non-negative stop-gradient contrastive evidence gate:
\begin{equation}
    w^{(g)}
    =
    \mathrm{sg}\!\left[
    \mathrm{clip}
    \left(1+a^{(g)}, w_{\min}, w_{\max}\right)
    \right].
\end{equation}
The clipping bounds keep the OPD update stable. The gate strengthens OPD for
rollouts whose tokens are better supported by the positive visual evidence view
and suppresses rollouts whose teacher support is not improved by that evidence.

The resulting gate assigns each rollout an evidence-dependent weight, connecting the teacher's visual comparison to the student's learning process. We incorporate this weight into the positive-view distillation objective below to translate trajectory-level evidence into token-level updates.

\subsection{V-Zero Objective}

After estimating the trajectory-level contrastive evidence gate, V-Zero discards the negative view from the training
target and distills only from the positive teacher view. At each student-induced
prefix, the positive-view local reverse-KL is
\begin{equation}
    D_{\mathrm{KL},+}^{(g,k)}
    =
    \sum_{v\in\mathcal{V}}
    \pi_s(v\mid x,y_{<k}^{(g)})
    \log
    \frac{
        \pi_s(v\mid x,y_{<k}^{(g)})
    }{
        \pi_t(v\mid x,z^{+},y_{<k}^{(g)})
    } .
\end{equation}
The underlying V-Zero distillation objective follows the standard reverse-KL
minimization convention:
\begin{equation}
    \label{eq:vzero-surrogate}
    \begin{aligned}
    \mathcal{L}_{\mathrm{V\mbox{-}Zero}}^{\mathrm{RKL}}(x,\mathcal{Y}(x))
    &=
    \frac{1}{G}\sum_{g=1}^{G}
    w^{(g)}\frac{1}{T_g}
    \sum_{k=1}^{T_g}
    D_{\mathrm{KL},+}^{(g,k)} .
    \end{aligned}
\end{equation}
In practice, sampled-token OPD forms the detached positive-view sampled
log-ratio score:
\begin{equation}
    \widehat{d}_{\mathrm{KL},+}^{(g,k)}
    =
    \mathrm{sg}\!\left[
    \log
    \frac{
        \pi_s(y_k^{(g)}\mid x,y_{<k}^{(g)})
    }{
        \pi_t(y_k^{(g)}\mid x,z^{+},y_{<k}^{(g)})
    }
    \right].
\end{equation}
The surrogate loss minimized in training is
\begin{equation}
    \begin{aligned}
    \widetilde{\mathcal{L}}_{\mathrm{V\mbox{-}Zero}}(x,\mathcal{Y}(x))
    &=
    \frac{1}{G}\sum_{g=1}^{G}
    w^{(g)}
    \frac{1}{T_g}
    \sum_{k=1}^{T_g}
    \widehat{d}_{\mathrm{KL},+}^{(g,k)}
    \\
    &\quad
    \log \pi_s(y_k^{(g)}\mid x,y_{<k}^{(g)}) .
    \end{aligned}
\label{surrogate}
\end{equation}
With $w^{(g)}$ and $\widehat{d}_{\mathrm{KL},+}^{(g,k)}$ detached, this surrogate gives the
contrastive-gated sampled reverse-KL gradient for the positive teacher view.
This formulation separates evidence comparison from token-level imitation: paired
visual evidence views decide how much to learn from each rollout, while the OPD
target remains the positive teacher distribution. In this way, V-Zero constructs dense
on-policy supervision without annotated textual answer labels and
without external reward signals. Algorithm 1 summarizes the complete training procedure.

\section{Experiments}
\begin{figure}
    \centering
    \includegraphics[width=1.0\linewidth]{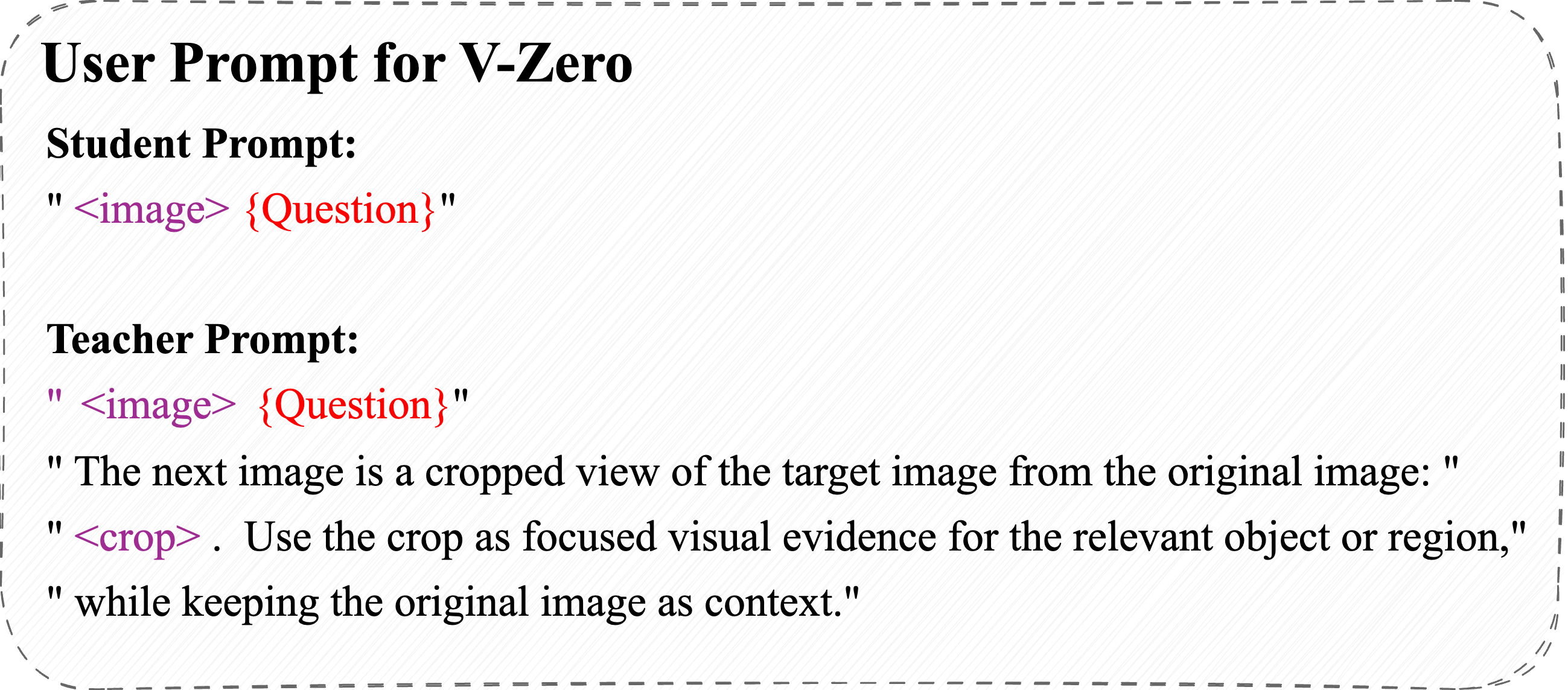}
    \caption{Prompt used in V-Zero. The placeholder \texttt{<image>}
    denotes the image-input, and \texttt{<crop>} denotes an
    additional teacher-side evidence image. Notice that these
    additional crops and instructions are not provided at
    inference time.}
    \label{fig:prompt}
\end{figure}

\begin{table*}[t]
    \centering
    \caption{
        Main results on fine-grained visual search and general multimodal
        perception benchmarks. V-Zero is compared with general
        large vision-language models and visually grounded reasoning models.
        \textsuperscript{*} denotes results obtained from our independent
        evaluation under the same experimental setting. Bold
        numbers denote the best results within each model group.
    }
    \label{tab:main-results}
    \small
    \setlength{\tabcolsep}{2.6pt}
    \renewcommand{\arraystretch}{0.95}

    \begin{tabular}{@{}l*{4}{c}|*{4}{c}|c@{}}
        \toprule
        \textbf{Method}
        & \multicolumn{4}{c|}{\textbf{Fine-Grained Visual Search}}
        & \multicolumn{4}{c|}{\textbf{General Benchmarks}}
        & \textbf{Overall} \\
        
        \cmidrule(lr){2-5}
        \cmidrule(lr){6-9}
        \cmidrule(l){10-10}

        & \textit{\hphantom{0}V*\hphantom{0}}
        & \textit{HR-4K}
        & \textit{HR-8K}
        & \textit{ZoomBench}
        & \textit{MME-RW}
        & \textit{MMStar}
        & \textit{MMVP}
        & \textit{BabyVision}
        & \textit{Avg.} \\
        \midrule

        \multicolumn{10}{c}{\textbf{General Large Vision-Language Models}} \\
        \midrule

        Qwen3-VL-4B\textsuperscript{*}~\cite{bai2025qwen3}
        & 81.7 & 78.5 & 75.3 & 40.4
        & 63.5 & 69.7 & 75.3 & 13.7 & 62.3 \\

        Qwen3.5-4B\textsuperscript{*}~\cite{qwen3.5}
        & 84.3 & 84.4 & 80.1 & 52.2
        & 69.2 & 71.8  & 79.7 & 19.1 & 67.6 \\

        Qwen3.5-9B\textsuperscript{*}
        & 89.0 & 87.8 & 84.5 & 56.8
        & 70.2 & 77.5  & 81.0 & 23.7 & 71.3 \\

        \midrule
        \multicolumn{10}{c}{\textbf{Visually Grounded Reasoning Models}} \\
        \midrule

        DeepEyes (7B) ~\cite{zheng2026deepeyesincentivizingthinkingimages}
        & 85.6 & 75.1 & 72.6 & 43.4
        & 64.1 & 63.5 & 72.0 & 12.9 & 61.1 \\

        Pixel-Reasoner (7B) ~\cite{wang2025pixelreasonerincentivizingpixelspace}
        & 84.3 & 72.6 & 66.1 & 43.8
        & 64.4 & 63.6 & 74.0  & 11.3 & 60.0 \\

        Thyme (7B)~\cite{zhang2025thymethinkimages}
        & 82.2 & 77.0 & 72.0 & 43.3
        & 64.8 & 64.2 & 73.3  & 13.7 & 61.3 \\

        DeepEyesV2 (7B)~\cite{hong2026deepeyesv2agenticmultimodalmodel}
        & 81.8 & 77.9 & 73.8 & 44.6
        & 64.9 & 63.9 & 72.0& 13.4 & 61.5 \\

        ZwZ-4B\textsuperscript{*}~\cite{wei2026zoomingzoomingregiontoimagedistillation}
        & \textbf{91.6} & 82.1 & 79.6 & 52.5
        & 68.5 & 71.1 & 78.3 & 16.2 & 67.5 \\

        ZwZ-8B\textsuperscript{*}~\cite{wei2026zoomingzoomingregiontoimagedistillation}
        & \textbf{91.6} & 84.9 & 82.4 & 56.6
        & 69.6 & 73.1 & 78.7 & 16.8 & 69.2 \\

        Vision-OPD-4B\textsuperscript{*}~\cite{yuan2026vision}
        & 86.4 & 82.4 & 77.9 & 57.6
        & 69.1 & 71.6 & 78.3 & 17.3 & 67.6 \\

        \textbf{V-Zero-4B (Ours)}
        & 89.8
        & \textbf{89.0}
        & \textbf{84.7}
        & \textbf{57.8}
        & \textbf{74.1}
        & \textbf{75.7}
        & \textbf{80.3}
        & \textbf{20.9}
        & \textbf{71.5} \\

        \bottomrule
    \end{tabular}
\end{table*}

\subsection{Experiment Setup}
\textbf{Baselines.}
We compare V-Zero with three groups of baselines. First, we evaluate Qwen3-VL and Qwen3.5 models to measure the gain over the backbone family~\cite{bai2025qwen3,qwen3.5}. Second, we compare with representative agentic visual reasoning and thinking-with-images systems, including DeepEyes \cite{zheng2026deepeyesincentivizingthinkingimages}, Thyme \cite{zhang2025thymethinkimages}, Pixel Reasoner \cite{wang2025pixelreasonerincentivizingpixelspace}, and DeepEyesV2 \cite{hong2026deepeyesv2agenticmultimodalmodel}. These systems enhance visual reasoning through agentic multimodal reasoning. Third, we compare with two distillation methods that transfer regional visual information to a student operating on the full image. ZwZ~\cite{wei2026zoomingzoomingregiontoimagedistillation} uses off-policy region-to-image distillation, while Vision-OPD~\cite{yuan2026vision} uses a crop-conditioned teacher to supervise the student on its own rollouts.

\noindent\textbf{Benchmarks.}
Following ZwZ \cite{wei2026zoomingzoomingregiontoimagedistillation}, we evaluate V-Zero on two groups of
benchmarks. The first group focuses on general perception in high-resolution or
fine-grained visual search scenarios, including HR-Bench~\cite{hrbench}, V* ~\cite{wu2024v}, and ZoomBench~\cite{wei2026zoomingzoomingregiontoimagedistillation}. The second group tests
general multimodal
understanding with MME-RealWorld~\cite{zhang2025mmerealworldmultimodalllmchallenge}, MMStar~\cite{chen2024we}, MMVP~\cite{tong2024eyes}, BabyVision~\cite{chen2026babyvision}.

\noindent\textbf{Training Dataset.}
We curate 5K high-quality training samples from LAION~\cite{schuhmann2022laion}, DocVQA~\cite{mathew2021docvqa}, ReasonSeg~\cite{lai2024lisa}, GQA~\cite{hudson2019gqanewdatasetrealworld}, TallyQA~\cite{acharya2018tallyqaansweringcomplexcounting}, TextVQA~\cite{singh2019vqamodelsread} and ChartQA~\cite{masry2022chartqa}. Each example contains a full image, a question, and a question-relevant regional crop annotated by Seed 2.1 Pro. See the Technical Supplement for annotation and crop-generation details. These crops are used only during training and are not provided at inference time. Notably, our method does not require additional tool-use trajectories or reasoning traces for cold-start initialization.

\noindent\textbf{Implementation Details.}
We use Qwen3.5-4B and Qwen3.5-27B as the default student and teacher,
respectively; additional settings are studied in the ablations. We implement
V-Zero with the VeRL framework~\cite{Sheng_2025}.
For optimization, we use a batch size of 32 and a mini-batch size of 16 with $G=8$ rollouts for each prompt. We configure maximum prompt
and response lengths of 25,000 and 2,048 tokens, respectively. We train with
a learning rate of $1\times 10^{-6}$. The distillation loss uses
the sampled-token reverse-KL estimator from VeRL's default OPD settings. The
contrastive evidence gating mechanism uses clipping bounds $w_{\min}=0$ and
$w_{\max}=2$. The student and teacher prompts are shown in Figure 4. More implementation details are provided in the Technical
Supplement.

\subsection{Main Results}
Table~\ref{tab:main-results} summarizes the results on four fine-grained visual search
benchmarks and four general multimodal perception benchmarks.
Compared with the Qwen3.5-4B backbone, V-Zero improves all eight
benchmarks, with gains of +5.5 on V*, +4.6 on HR-4K, +4.6 on HR-8K, +5.6 on ZoomBench, +4.9 on MME-RW, +3.9 on MMStar, +0.6 on MMVP, and +1.8 on BabyVision. The average across the four fine-grained benchmarks increases from 75.3 to 80.3, and the overall average across all eight benchmarks increases from 67.6 to 71.5. The consistent
gains on the general benchmarks indicate that V-Zero
strengthens localized visual reasoning without sacrificing the
general multimodal capabilities of the backbone.  The qualitative attention maps in Figure~\ref{fig:placeholder-visualization} further show that V-Zero focuses more strongly on task-relevant visual regions.

The comparison with Qwen3.5-9B further illustrates where the gains are concentrated. V-Zero-4B outperforms this larger model on all four fine-grained benchmarks, with gains of +0.8 on V*, +1.2 on HR-4K, +0.2 on HR-8K, and +1.0 on ZoomBench, as well as +3.9 on MME-RW. The average across the four fine-grained benchmarks increases from 79.5 to 80.3, and the overall average across all eight benchmarks increases from 71.3 to 71.5. These results are consistent with the training objective's emphasis on task-relevant visual detail.

\noindent\textbf{Training Cost.}
V-Zero-4B trains in 7.3 hours on eight H100 GPUs. As shown in Figure~\ref{fig:paradigms}, it requires 1.83 hours per billion parameters, compared with 6.00 for ZwZ-4B and 6.86 for DeepEyes-7B, while achieving higher overall accuracy than both. Compared with Vision-OPD-4B, V-Zero-4B requires only one additional teacher forward pass per training rollout to score the same trajectory under the negative visual view. With this modest overhead of about 11\% in training time per parameter, V-Zero-4B improves average accuracy across the eight benchmarks from 67.6\% to 71.5\%, a gain of 3.9 percentage points.

\subsection{Ablation Study}

\noindent\textbf{OPD variants.} We compare three OPD variants to isolate the gains from positive-view supervision and trajectory-level gating. As shown in Table~\ref{tab:ablation-epg}, conditioning the teacher on the positive view improves the perception average from 77.1 to 78.6, consistent with the benefit of crop-conditioned supervision reported by Vision-OPD~\cite{yuan2026vision}. As discussed in our analysis of OPD, the teacher still supervises each next token without explicitly assessing whether the visual evidence supports the rollout as a whole. V-Zero supplies this trajectory-level signal by comparing the teacher's likelihood of the same rollout under the positive and negative views. Weighting the distillation update by this contrast raises the average to 80.3, a further 1.7 points over positive view OPD. The gains extend across all four benchmarks, with improvements of +1.3 on V*, +2.4 on HR-4K, +1.9 on HR-8K, and +1.5 on ZoomBench over positive-view OPD. Together, these results highlight the complementary roles of regional supervision and evidence gating: the positive view provides a focused teacher target, while the gate directs stronger updates toward trajectories with greater relative visual support.

\begin{table}[H]
\centering
\caption{Ablation of OPD variants.}
\label{tab:ablation-epg}
\small
\setlength{\tabcolsep}{2.1pt}
\renewcommand{\arraystretch}{0.9}
\resizebox{\columnwidth}{!}{%
\begin{tabular}{@{}c|ccccc@{}}
\toprule
\textbf{Method} & \textbf{\hphantom{0}V*\hphantom{0}} & \textbf{HR-4K} & \textbf{HR-8K} & \textbf{ZoomBench} & \textbf{Avg} \\
\midrule
Standard OPD & 87.4 & 86.0 & 82.3 & 52.7 & 77.1 \\
Positive view OPD & 88.5 & 86.6 & 82.8 & 56.3 & 78.6 \\
\textbf{V-Zero} & \textbf{89.8} & \textbf{89.0} & \textbf{84.7} & \textbf{57.8} & \textbf{80.3} \\
\bottomrule
\end{tabular}}
\end{table}

\noindent\textbf{Choice of visual evidence.} We vary the visual evidence supplied to the teacher to test how the choice of positive and negative views affects fine-grained accuracy. Table~\ref{tab:ablation-negative-view} suggests that the choice of negative view matters through its contrast with the task-relevant positive view. A blank view removes the relevant detail entirely and yields 78.4. A nearby crop yields a lower score 75.4. We attribute this drop to the visual cues shared by the nearby crop and the positive view. These cues weaken the contrast between the two views when the teacher scores the same rollout. A random crop achieves 80.3 and outperforms the blank view on all four benchmarks. We attribute this 1.9-point gain over the blank view to the more robust training signal that random crops provide by exposing the gate to diverse visual distractors. Randomizing the positive view as well lowers the average to 72.5, suggesting that the contrast is most useful when the positive view contains evidence for the task.

\begin{table}[H]
\centering
\caption{Ablation of visual evidence selection. Pos. and Neg. denote the positive and negative views used for gating, respectively.}
\label{tab:ablation-negative-view}
\small
\setlength{\tabcolsep}{2.1pt}
\renewcommand{\arraystretch}{0.9}
\resizebox{\columnwidth}{!}{%
\begin{tabular}{@{}cc|ccccc@{}}
\toprule
\textbf{Pos.} & \textbf{Neg.} & \textbf{\hphantom{0}V*\hphantom{0}} & \textbf{HR-4K} & \textbf{HR-8K} & \textbf{ZoomBench} & \textbf{Avg} \\
\midrule
Random & Random & 83.3 & 82.4 & 77.3 & 47.2 & 72.5 \\
Task-relevant & Blank & 86.9 & 88.3 & 82.9 & 55.3 & 78.4 \\
Task-relevant & Nearby & 84.3 & 84.1 & 80.1 & 52.9 & 75.4 \\
\textbf{Task-relevant} & \textbf{Random} & \textbf{89.8} & \textbf{89.0} & \textbf{84.7} & \textbf{57.8} & \textbf{80.3} \\
\bottomrule
\end{tabular}}
\end{table}
\noindent\textbf{Effect of teacher size.} We compare 9B and 27B teachers to test whether a larger teacher improves the performance of the same 4B student.
As shown in Table~\ref{tab:ablation-size}, increasing the teacher size from 9B to 27B improves the perception average from 78.6 to 80.3. The larger teacher improves HR-4K by 2.9 points, HR-8K by 2.3 points, and ZoomBench by 1.9 points. V* changes little, from 90.0 to 89.8.

\begin{table}[H]
    \centering
    \caption{Ablation of teacher model sizes.}
    \label{tab:ablation-size}
    {\small
    \setlength{\tabcolsep}{2.1pt}
    \renewcommand{\arraystretch}{0.9}
    \resizebox{\columnwidth}{!}{%
    \begin{tabular}{@{}cc|ccccc@{}}
        \toprule
        \textbf{Teacher} & \textbf{Student} & \textbf{\hphantom{0}V*\hphantom{0}} & \textbf{HR-4K} & \textbf{HR-8K} & \textbf{ZoomBench} & \textbf{Avg} \\
        \midrule
        % 4B & 4B \\
        9B & 4B & \textbf{90.0} & 86.1 & 82.4 & 55.9 & 78.6 \\
        27B & 4B & 89.8 & \textbf{89.0} & \textbf{84.7} & \textbf{57.8} & \textbf{80.3} \\
        \bottomrule
    \end{tabular}
    }
    }
\end{table}

\noindent\textbf{Rollout group size.}
We vary the number of rollouts per prompt to test whether a larger comparison group improves fine-grained accuracy in Table~\ref{tab:ablation-rollouts}. Increasing the group size
from $G=2$ to $G=4$ raises the perception average from 73.4 to 77.9, and further
increasing it to $G=8$ achieves 80.3. Performance improves consistently across
all four benchmarks as the group size increases. This indicates that a larger rollout group provides more diverse
within-prompt comparisons, enabling the gate to distinguish
visually well-supported trajectories more reliably.

\begin{table}[H]
    \centering
    \caption{Ablation of rollout group size.}
    \label{tab:ablation-rollouts}
    {\small
    \setlength{\tabcolsep}{2.9pt}
    \renewcommand{\arraystretch}{0.9}
    \begin{tabular}{@{}c|ccccc@{}}
        \toprule
        \textbf{Rollouts} & \textbf{\hphantom{0}V*\hphantom{0}} & \textbf{HR-4K} & \textbf{HR-8K} & \textbf{ZoomBench} & \textbf{Avg} \\
        \midrule
        $G=2$ & 84.3 & 83.1 & 75.9& 50.1 & 73.4 \\
        $G=4$ & 87.4 & 85.5 & 83.3 & 55.3 & 77.9 \\
        $G=8$ & \textbf{89.8} & \textbf{89.0} & \textbf{84.7} & \textbf{57.8} & \textbf{80.3} \\
        \bottomrule
    \end{tabular}
    }
\end{table}

% \begin{figure}
%     \centering
%     \includegraphics[width=1\linewidth]{curve.png}
%     \caption{Enter Caption}
%     \label{fig:placeholder}
% \end{figure}
\begin{figure}[t]
    \centering
    \includegraphics[width=1\linewidth]{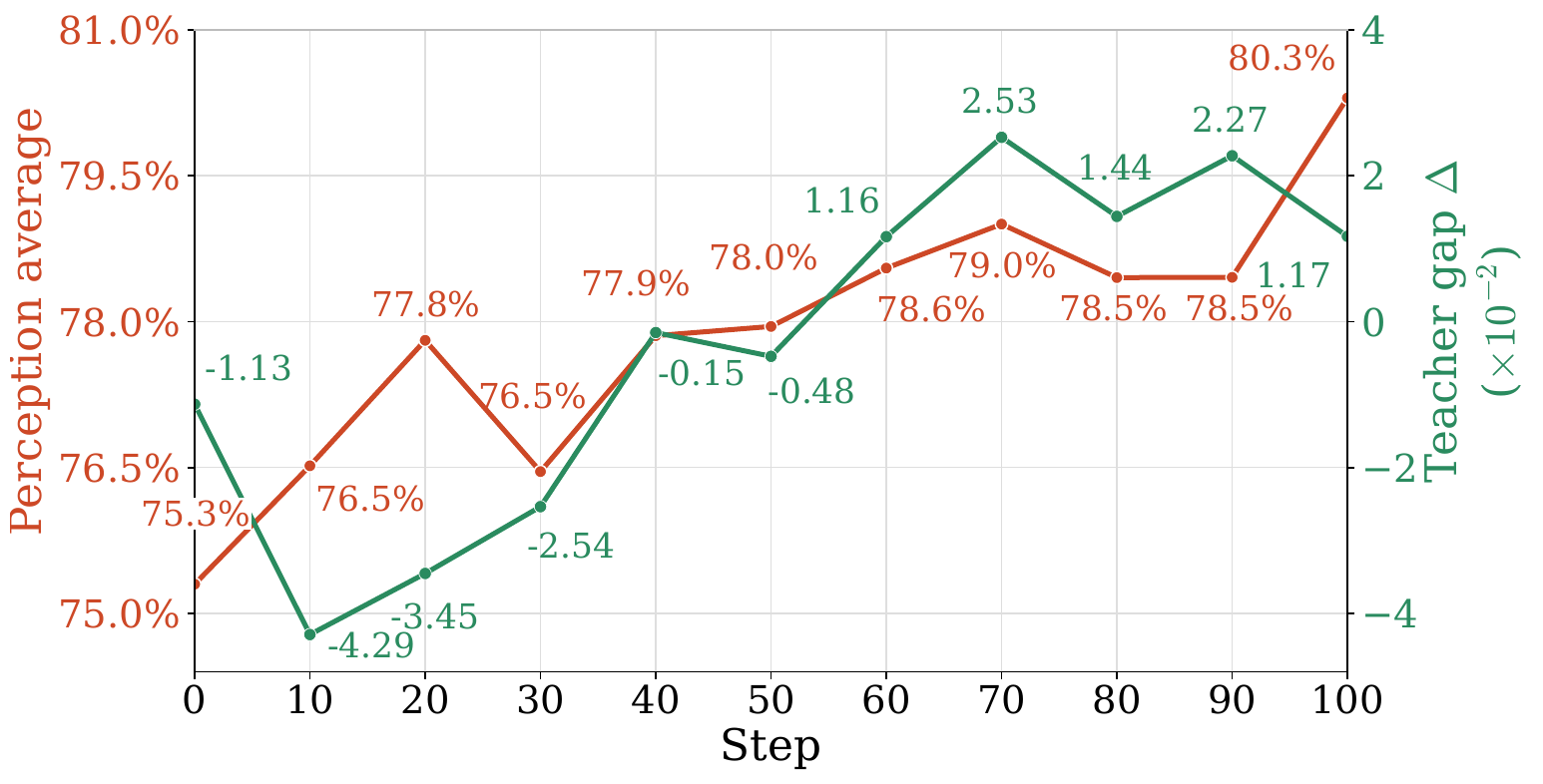}
    \caption{Training dynamics of the perception average and
    the mean positive-minus-negative teacher log-likelihood gap $\Delta$. Step 0 denotes the
    Qwen3.5-4B backbone before training.}
    \label{fig:dynamics}
\end{figure}
% \begin{table}[t]
%     \centering
%     {\small
%     \setlength{\tabcolsep}{2.7pt}
%     \renewcommand{\arraystretch}{0.9}
%     \begin{tabular}{@{}lcccccc@{}}
%         \toprule
%         \textbf{Step} & \textbf{0} & \textbf{30} & \textbf{40} & \textbf{50} & \textbf{60} & \textbf{70}  \textbf{80} \\
%         \midrule
%         V* & 84.3 & 88.0 & 86.9 & 87.4 & 89.0 & 89.0 & \textbf{89.5}\\
%         HR-4K & 84.4 & 86.9 & 85.8 & 86.8 & 88. & 85.6 \\
%         HR-8K & 80.1 & 81.7 & 81.6 & \textbf{83.0} & 82.1 & 82.0 \\
%         ZoomBench & 52.2 & 55.2 & 53.5 & 56.7 & \textbf{57.7} & 55.6 \\
%         \textbf{Perc. Avg.} & 75.3 & 77.2 & 77.4 & 78.4 & \textbf{79.2} & 77.8 \\
%         \bottomrule
%     \end{tabular}
%     }
%     \caption{Ablation of training steps. Step 0 denotes the Qwen3.5-4B base model before V-Zero training. Perception Avg. is computed over V*, HR-4K, HR-8K, and ZoomBench.}
%     \label{tab:ablation-steps}
% \end{table}
\noindent\textbf{Training dynamics.}
Figure~\ref{fig:dynamics} tracks the student's average perception accuracy and the teacher's log-likelihood gap $\Delta$ between the positive and negative views. The teacher gap peaks at step 70, while the student's average perception accuracy rises from 75.3 at step 0 to a peak of 80.3 at step 100. This pattern suggests that the student learns from the trajectory-level contrast conveyed by the gate.

% Different from these works, we
% study OPD for fine-grained visual reasoning through a negative-free
% stop-gradient alignment view and convert teacher-side evidence comparisons under paired
% positive and negative visual evidence views into trajectory-level contrastive
% evidence gates.

\section{Conclusion}
We introduced V-Zero, an efficient post-training framework that equips on-policy distillation with trajectory-level discrimination for fine-grained visual reasoning. By viewing standard OPD as a stop-gradient alignment objective, we reveal its lack of an explicit mechanism for distinguishing visually grounded trajectories from locally plausible but drifting ones. V-Zero addresses this limitation by contrasting teacher support under relevant and irrelevant visual evidence and converting the resulting group-relative signal into a gate for positive-view distillation. Without annotated reasoning traces, external reward signals, or additional inference-time operations, V-Zero improves the Qwen3.5-4B backbone by 3.9 points on average while achieving 3× and 6× training speedups over representative SFT and RL baselines. These results demonstrate that contrastive visual evidence can turn dense token-level supervision into selective trajectory-level learning, offering a simple approach to efficient multimodal post-training. We hope future work can explore automatically discovered evidence and extend trajectory-level gating to broader multimodal scenarios.

\section*{LLM Usage Statement}
We used GPT-5.6 to assist with manuscript polishing, generate preliminary figure sketches that were subsequently redesigned by us, and help develop the experimental code. The authors take full responsibility for the manuscript, experimental code, results, and conclusions.

\bibliography{aaai2027}

\end{document}